\documentclass{article}

    \PassOptionsToPackage{numbers, sort&compress}{natbib}
\usepackage[preprint]{neurips_2026}

\usepackage[utf8]{inputenc} 
\usepackage[T1]{fontenc}    
\usepackage{hyperref}       
\usepackage{url}            
\usepackage{booktabs}       
\usepackage{amsfonts}       
\usepackage{nicefrac}       
\usepackage{microtype}      
\usepackage{xcolor}         

\usepackage{graphicx}
\usepackage{amsmath}
\usepackage{multirow}
\usepackage{amssymb}
\usepackage{colortbl}
\usepackage{wrapfig}
\usepackage[most]{tcolorbox}
\usepackage{makecell}
\usepackage{xspace}
\usepackage{subcaption}
\usepackage{enumitem}
\usepackage{pifont}
\usepackage{color, colortbl}
\usepackage{algorithm}
\usepackage{algorithmicx}
\usepackage{algpseudocode}
\usepackage[nameinlink,noabbrev]{cleveref}
\usepackage{tabularx}
\usepackage[font=small]{caption}

\newcommand{\kitti}{KITTI\xspace}
\newcommand{\nyu}{NYU-v2\xspace}

\newcommand{\ddad}{DDAD\xspace}
\newcommand{\aTwoDTwo}{A2D2\xspace}
\newcommand{\taskonomy}{Taskonomy\xspace}
\newcommand{\ibims}{IBims-1\xspace}
\newcommand{\hmThreeD}{HM3D\xspace}
\newcommand{\scannetpp}{ScanNet++\xspace}
\newcommand{\mThreeD}{Matterport3D\xspace}
\newcommand{\panoGVTwo}{Pano3D-GV2\xspace}

\newcommand{\absRel}{A.Rel\xspace}
\newcommand{\rmse}{RMSE\xspace}

\newcommand{\ours}{DepthAgent\xspace}

\newcommand{\unidac}{UniDAC\xspace}
\newcommand{\uniKThreeD}{UniK3D\xspace}
\newcommand{\unidepth}{UniDepth\xspace}
\newcommand{\metricThreeD}{Metric3D\xspace}
\newcommand{\metricThreeDvTwo}{Metric3Dv2\xspace}

\newcounter{takeawayonly}

\newcounter{conclusion}

\renewcommand{\paragraph}[1]{\vspace{1.25mm}\noindent\textbf{#1}}

\makeatletter
\DeclareRobustCommand\onedot{\futurelet\@let@token\@onedot}
\def\@onedot{\ifx\@let@token.\else.\null\fi\xspace}

\def\eg{\emph{e.g}\onedot} 

\def\ie{\emph{i.e}\onedot}

\makeatother

\definecolor{green}{HTML}{0aa344}
\definecolor{red}{HTML}{c93756}
\definecolor{darkgreen}{HTML}{068C52}
\definecolor{fullgreen}{rgb}{0.502, 0.788, 0.643}
\definecolor{fullred}{rgb}{0.800, 0.447, 0.541}
\definecolor{lightgreen}{RGB}{225, 239, 217}
\definecolor{lightblue}{RGB}{203, 220, 235}
\definecolor{fullgray}{RGB}{219, 223, 234}
\definecolor{fullpurple}{RGB}{205, 193, 255}
\definecolor{darkred}{RGB}{204, 114, 138}
\definecolor{darkpurple}{RGB}{171, 151, 255}
\definecolor{darkgray}{RGB}{114, 114, 114}

\definecolor{teal}{HTML}{14B8A6}
\definecolor{sky}{HTML}{38BDF8}
\definecolor{indigo}{HTML}{6366F1}
\definecolor{navy}{HTML}{1E3A8A}

\definecolor{amber}{HTML}{F59E0B}
\definecolor{coral}{HTML}{FF6B6B}
\definecolor{peach}{HTML}{FFB4A2}

\definecolor{sage}{HTML}{A8C3A1}
\definecolor{dustyblue}{HTML}{9BB4C7}
\definecolor{mauve}{HTML}{BFA6C9}
\definecolor{clay}{HTML}{C9B29B}

\usepackage{microtype}

\title{DepthAgent: Towards Better Universal Depth Estimation via Sample-wise Expert Selection}

\author{
Jie Zhu\textsuperscript{1}\quad Girish Chandar Ganesan\textsuperscript{1}\quad Xiaoming Liu\textsuperscript{1,2} \\
\textsuperscript{1}Michigan State University \quad \textsuperscript{2}University of North Carolina at Chapel Hill\\
{\tt\small \{zhujie4, ganesang\}@msu.edu \quad liuxm@cs.unc.edu } \\
}

\begin{document}

\maketitle

\begin{abstract}
Monocular metric depth estimation has achieved strong progress with large-scale training and universal-camera modeling, yet robust deployment across diverse camera settings, such as perspective, fisheye, and panoramic images, remains challenging. Existing methods typically rely on a single depth estimator, overlooking that different models encode different camera assumptions and perform best under different input domains. In this paper, we show that depth experts exhibit strong sample-wise complementarity: model preference is highly correlated with camera geometry, and multi-model fusion brings the largest gains on difficult samples where individual experts are unreliable. Motivated by these observations, we propose \textbf{\ours}, a vision-language agent for adaptive monocular depth estimation. DepthAgent treats existing depth models as frozen tools and learns to analyze scene and camera cues, invoke suitable experts through multi-turn tool utilization, and select or fuse their predictions for each input. To optimize such discrete decision-making toward dense geometric quality, we design a multi-reward reinforcement fine-tuning scheme that jointly encourages valid tool execution, camera/scene analysis, expert-selection quality, and inference efficiency. Extensive experiments across perspective, fisheye, and panoramic benchmarks show that \ours consistently outperforms individual experts, fixed model fusion, and different selection strategies, with strong improvements on challenging samples, highlighting the critical role of expert selection and fusion. The code and model will be released upon publication.
\end{abstract}
\section{Introduction} \label{sec:intro}

Monocular depth estimation is a fundamental problem for 3D scene understanding, with broad applications in autonomous driving, robotics, and augmented/virtual reality~\cite{wang2019pseudo,park2021pseudo,dong2022towards,du2020depthlab}. Recent advances have substantially improved monocular depth prediction through large-scale supervised and self-supervised learning~\cite{ranftl2020towards,yang2024depth,yang2024depthv2}, diffusion-based geometric priors~\cite{ke2024repurposing,rombach2022high}, and zero-shot metric depth estimation~\cite{yin2023metric3d,hu2024metric3d,piccinelli2024unidepth}. Despite this progress, many real-world applications require metric depth, where accurate scale is critical for reliable spatial reasoning and visual grounding~\cite{kumar2022deviant, kumar2025charm3r, zhang2025unleashing, zhang2026towards}. This necessity has driven growing interest in monocular metric depth estimation (MMDE), which aims to recover depth in absolute units rather than relative ordering. However, real-world deployment rarely follows a single imaging assumption. Images may come from standard perspective cameras, challenging perspective settings, fisheye lenses, or equirectangular projected (ERP) panoramas. Although recent works attempt to extend monocular depth estimation to large-FoV and universal-camera settings~\cite{jiang2021unifuse,conf/cvpr/LiGY0DR22/omnifusion,conf/eccv/ShenLLNZZ22/panoformer,conf/iccv/YunSLLR23/egformer,conf/cvpr/AiCCSW23/hrdfuse,guo2025depth,ganesan2026unidac,piccinelli2025unik3d}, robust depth estimation across heterogeneous camera geometries and diverse scenarios remains challenging.

\begin{figure}[t!]
  \centering
  \includegraphics[width=\linewidth]{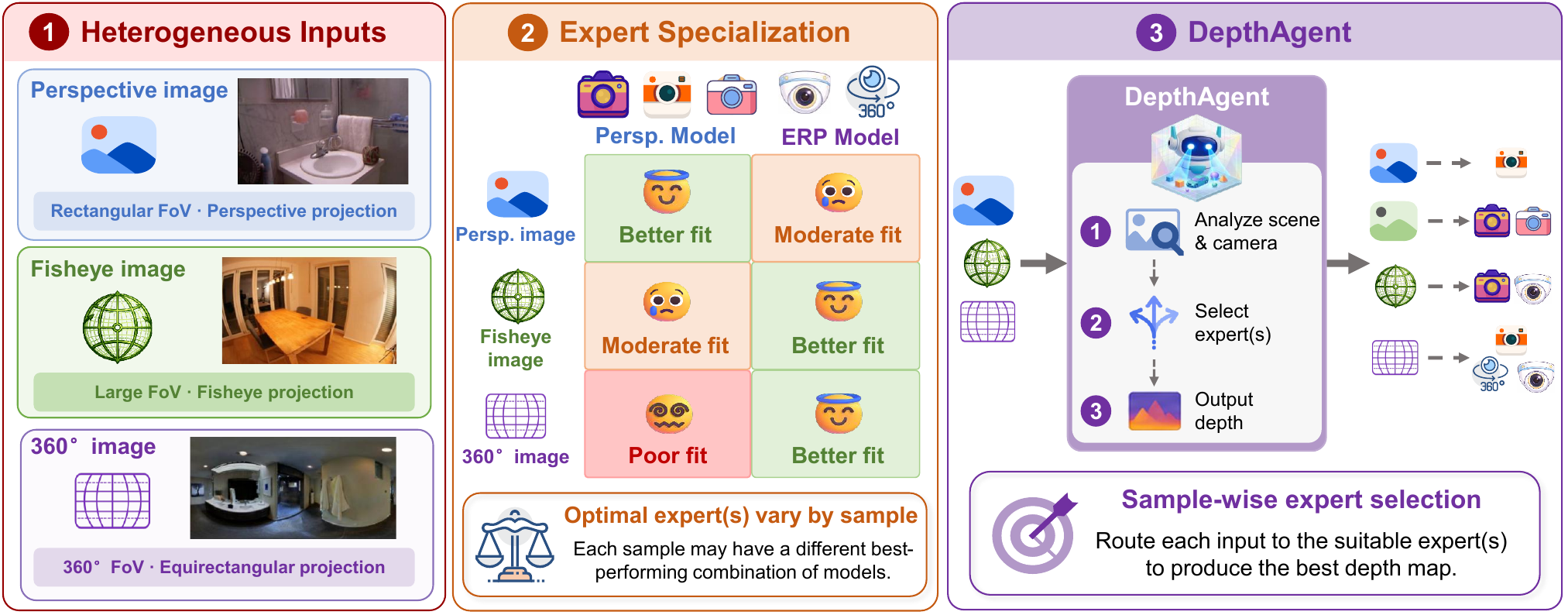}
  \caption{
    \textbf{Motivation of \ours.} Real-world inputs span heterogeneous camera domains, including perspective, fisheye, and panoramic images, for which different depth experts exhibit different strengths. Instead of relying on a single model for all inputs, \ours uses scene and camera cues to select suitable expert(s) on a per-sample basis, producing more reliable depth maps across diverse camera settings.
  }
  \label{fig:teaser}
  \vspace{-2em}
\end{figure}

A central challenge in robust monocular depth estimation lies not only in the lack of strong depth models, but also in the fact that existing models encode different imaging assumptions and thus perform best under different camera geometries or data distributions. For instance, perspective-oriented models are well suited to standard perspective images, while ERP-trained models are better adapted to large-FoV and native panoramic/360$^\circ$ inputs. As a result, no single model is consistently optimal across scenarios, motivating adaptive model selection or combination.
However, model combination remains underexplored in monocular depth estimation. Although routing and expert-selection strategies have been studied in language, vision, and multimodal systems~\cite{chen2023frugalgpt,ong2024routellm,lu2024routing,zong2024mova,ye2023taskexpert,zhu2026fusionagent}, they cannot be directly applied to depth estimation, where success depends on dense geometric accuracy rather than discrete labels or high-level preferences. A naive solution is to select a single fixed model or to fuse all expert predictions using a predefined rule~\cite{liu2025person}, but both are limited: the former overlooks complementary expert strengths, whereas the latter may indiscriminately incorporate unreliable predictions and degrade the final depth estimate. Therefore, the key missing component is an input-adaptive mechanism that determines whether experts should be combined and which experts should participate. This leads to a research question: \textit{given multiple depth models for the same task, how should one choose the most suitable model or model combination for each test sample?}

To examine whether such a mechanism is necessary and feasible, we first conduct a systematic analysis of single-model performance, oracle-style model selection, and multi-model fusion across diverse models and datasets. Our analysis reveals that \textbf{multi-model fusion often outperforms individual models} on most samples, and model preference is strongly correlated with camera geometry: perspective-oriented models are generally more effective on standard perspective inputs, while ERP-trained models are more reliable for native panoramic images. These highlight the potential benefit of exploiting complementary strengths among depth experts. We further observe that the benefit of fusion is not uniform, but becomes more pronounced on difficult samples where single-model predictions are less reliable. These findings suggest that robust monocular depth estimation should not rely on a fixed estimator or a static fusion rule, but instead requires an input-adaptive mechanism that can decide which experts to use and when to combine them.

Motivated by these observations, we recast robust universal depth estimation as an input-adaptive solution selection problem. Instead of seeking a single estimator or applying a fixed fusion rule, the goal is to decide for each input which depth expert(s) to trust, whether fusion is beneficial, and how much computation is justified. We propose \textbf{\ours}, an agentic framework for adaptive monocular depth estimation. 
\ours treats existing depth models as frozen tools and uses a Vision-Language Model (VLM) agent as a learned controller for sample-wise expert selection. The agent reasons over scene and camera cues, performs multi-turn tool use, and finally selects or fuses expert predictions, which are difficult to capture with a fixed hand-crafted router. Since the desired behavior involves discrete expert selection while the final objective is dense geometric quality, we optimize the agent with reinforcement fine-tuning based on GRPO~\cite{shao2024deepseekmath}, avoiding fixed oracle trajectories while enabling dynamic control over the performance-efficiency trade-off.
We further design a \textbf{multi-reward scheme} that jointly encourages valid tool execution, camera- and scene-aware reasoning, empirically strong expert selection, and computation-aware efficiency. Experiments across perspective, fisheye, and panoramic benchmarks show consistent gains over individual experts, static fusion baselines, and different expert-selection strategies, especially on challenging samples where individual predictions are often unreliable. Our contributions are summarized as follows:
\begin{itemize}[noitemsep, topsep=2pt, leftmargin=1em]
    \item We conduct a systematic analysis of single-model selection and multi-model fusion for monocular depth estimation across perspective and panoramic scenarios, revealing strong model complementarity and camera-dependent expert preference.
    \item We formulate robust universal depth estimation as an input-adaptive solution selection problem and propose \ours, a VLM-based agent that adaptively selects depth experts and fusion strategies for each sample.
    \item We design a multi-reward reinforcement fine-tuning scheme that encourages valid tool execution, camera- and scene-aware decision-making, and quality-efficiency balancing.
    \item We validate \ours on diverse depth benchmarks, showing consistent improvements over individual experts, non-agentic fusion baselines, and different selection strategies, with especially strong gains on hard samples.
\end{itemize}

\begin{table}[t!]
  \centering
  \caption{
    \textbf{Model-family preference across camera-domain groups.}
    Best-single columns indicate the percentage of samples for which a Perspective or ERP model achieves the highest single-model performance. Oracle-presence columns indicate the percentage of samples in which at least one model from the corresponding family is selected by the oracle solution. Avg. gain denotes the average performance improvement achieved by fusion over the corresponding best single model across the evaluated perspective and ERP models.
  }
  \vspace{0.5em}
  \label{tab:cross_domain}
  \setlength{\tabcolsep}{6pt}
  \renewcommand{\arraystretch}{1.12}
  \resizebox{0.75\textwidth}{!}{
  \begin{tabular}{lccccc}
    \toprule
    \multirow{2}{*}{\textbf{Dataset group}}
      & \multicolumn{2}{c}{\textbf{Best single model}}
      & \multicolumn{2}{c}{\textbf{Oracle solution presence}}
      & \multirow{2}{*}{\textbf{Avg. gain}} \\
    \cmidrule(lr){2-3}
    \cmidrule(lr){4-5}
      & Perspective & ERP
      & Perspective & ERP
      & $\Delta \delta_1$ \\
    \midrule
    Perspective & 80.1 & 19.9 & 99.0 & 58.8 & 0.019 \\
    ERP variant & 77.1 & 22.9 & 94.8 & 52.1 & 0.016 \\
    Native ERP  &  2.2 & 97.8 & 36.5 & 98.2 & 0.044 \\
    \bottomrule
  \end{tabular}
  }
  \vspace{-1em}
\end{table}

\section{Related Works}
\label{sec:related_works}

\paragraph{Monocular Depth Estimation.}
Early monocular depth estimation primarily focused on perspective images, improving generalization with large-scale labeled datasets~\cite{ranftl2020towards, yin2021learning}, unlabeled data~\cite{yang2024depth, yang2024depthv2}, and generative priors from pre-trained diffusion models~\cite{ke2024repurposing, rombach2022high}. Recent zero-shot metric depth methods further improve scale recovery by exploiting camera parameters through explicit supervision or canonical parameterization~\cite{guizilini2023towards, hu2024metric3d, yin2023metric3d, piccinelli2024unidepth, piccinelli2025unidepthv2}. Beyond perspective images, large-FoV inputs such as fisheye and 360$^\circ$ images have been studied to handle richer context and severe geometric distortions~\cite{jiang2021unifuse, conf/cvpr/LiGY0DR22/omnifusion, conf/eccv/ShenLLNZZ22/panoformer, conf/iccv/YunSLLR23/egformer, conf/cvpr/AiCCSW23/hrdfuse}. Yet, due to limited large-FoV depth data and strong in-domain assumptions, existing models still struggle to generalize across camera types, even with recent efforts toward distortion modeling and unified cross-camera representations~\cite{conf/nips/SuG17/deformcnn360, zhu2019deformableconvnetsv2deformable, xiong2024efficient, journals/ral/JiangSZDH21/unifuse, rey2022360monodepth, feng2023simfirsimpleframeworkfisheye, guo2025depth, ganesan2026unidac, piccinelli2025unik3d}.
This highlights the importance of input-adaptive expert selection for robust real-world deployment, motivating \ours to choose the most suitable expert(s) for each input.

\paragraph{Model Fusion and Expert Selection.}
Expert fusion and dynamic expert selection have been widely studied in language, vision, and multimodal  systems~\cite{ong2024routellm, chen2023atm, lu2024routing, chen2023frugalgpt, wu2025tove, zong2024mova,liu2025person, ye2023taskexpert, li2025simmlm,zuo20254kagent, guo2026holistic, zhu2025quality, zhu2026fusionagent}. However, these ideas remain underexplored in depth estimation, with only limited studies on prediction fusion or complementary cues~\cite{ren2022adaptive, xia2020generating, dai2023multi}, while DepthFusion~\cite{obukhov2025fourth} indicates the potential of variance-aware expert selection. In contrast, we systematically study sample-wise model selection and fusion for depth estimation, demonstrating that adaptive expert choice and fusion can consistently improve performance, especially on challenging samples.

\paragraph{Agentic Tool Use and Reinforcement Fine-tuning.}
Recent studies have advanced multimodal models into agentic systems that can plan, invoke tools, and iteratively refine predictions for complex visual tasks~\cite{zhou2025revpt,liu2025longvideoagent,chen2025lvagent,zuo20254kagent,zhu2026fusionagent}. Meanwhile, Reinforcement Learning (RL) has become an effective paradigm for improving the reasoning and problem-solving abilities of LLMs and VLMs~\cite{sutton1998reinforcement,kaelbling1996reinforcement,zhu2024fairness,matsuo2022deep,han2023survey,guo2025deepseek,bai2022training,sun2023aligning,shao2024deepseekmath,hui2024qwen2}. Recent works further adapt GRPO and its variants~\cite{guo2025deepseek,yu2025dapo,liu2025understanding,zheng2025group,liu2026gdpo} to multimodal tasks with rule-based rewards~\cite{shen2025vlm,huang2025vision,liu2025visual,tan2025reason,zhang2025r1,zhu2026can,chen2025suitability,peng2025lmm,yang2025r1,zhu2026fusionagent}. In contrast, we design selection, efficiency, and scene-awareness rewards to jointly optimize the effectiveness and efficiency of \ours for depth estimation.
\section{Method Overview} \label{sec:methods}

We first conduct fusion analysis to understand when different depth experts or fusion solutions succeed or fail. These findings motivate the rest of our design. Based on them, we introduce \ours, which learns to dynamically select suitable solutions for each input image. Finally, we derive the reward functions from the same empirical observations, encouraging the agent to favor solutions that yield better depth quality across different scenarios.

\subsection{Fusion Analysis} \label{subsec:fusion_analysis}

\paragraph{Analysis setup.}
We first specify the candidate experts, fusion strategies, and evaluation protocol used in our fusion analysis. We consider perspective-oriented experts, including \unidepth~\cite{piccinelli2024unidepth}, \metricThreeD~\cite{yin2023metric3d}, and \metricThreeDvTwo~\cite{hu2024metric3d}, and ERP-trained experts, including \unidac~\cite{ganesan2026unidac} and \uniKThreeD~\cite{piccinelli2025unik3d}. Besides individual experts, we evaluate pixel-wise mean, max, and min fusion~\cite{zhang2020ifcnn, ma2022swinfusion}, as well as DepthFusion~\cite{obukhov2025fourth}. For each sample, a \emph{solution} refers to either a single expert or a combination of multiple experts with a fusion strategy. \emph{Oracle solution} is defined as the candidate solution that achieves the best $\delta_1$, providing an empirical upper bound for sample-wise selection and fusion. We conduct the analysis on perspective datasets, including \aTwoDTwo~\cite{geyer2020a2d2}, \ddad~\cite{Guizilini2020ddad}, \hmThreeD~\cite{ramakrishnan2021habitat}, and \taskonomy~\cite{zamir2018taskonomy}, together with their ERP variants, constructed by transforming images into ERP patches, since ERP data are relatively scarce (details in Appendix). To cover native ERP inputs, we include \scannetpp~\cite{yeshwanth2023scannet++}, \mThreeD~\cite{chang2017matterport3d}, and \panoGVTwo~\cite{albanis2021pano3d}. For each dataset, we randomly select 200 training samples for analysis, ensuring that no test-set bias is introduced.

\paragraph{Camera geometry is the primary factor in model preference.}
Tab.~\ref{tab:cross_domain} shows that model preference is strongly tied to camera geometry. Perspective models dominate perspective datasets, winning 80.1\% of samples, and remain preferred on ERP variants. This suggests that ERP reprojection of perspective images does not fully reproduce the distortions and model preferences of native panoramic data. In contrast, native ERP datasets show a clear reversal, where ERP-trained models win 97.8\% of samples. Oracle solutions further reveal cross-family complementarity, with the dominant model family varying consistently with the input camera type. These results suggest a clear camera-dependent preference: even though ERP models can process perspective-like inputs, perspective-trained models remain superior on perspective data, whereas ERP-trained models perform better on native ERP images. Therefore, \textit{model selection should depend on the camera type.}

\begin{figure}[t!]
  \centering
  \includegraphics[width=.95\linewidth]{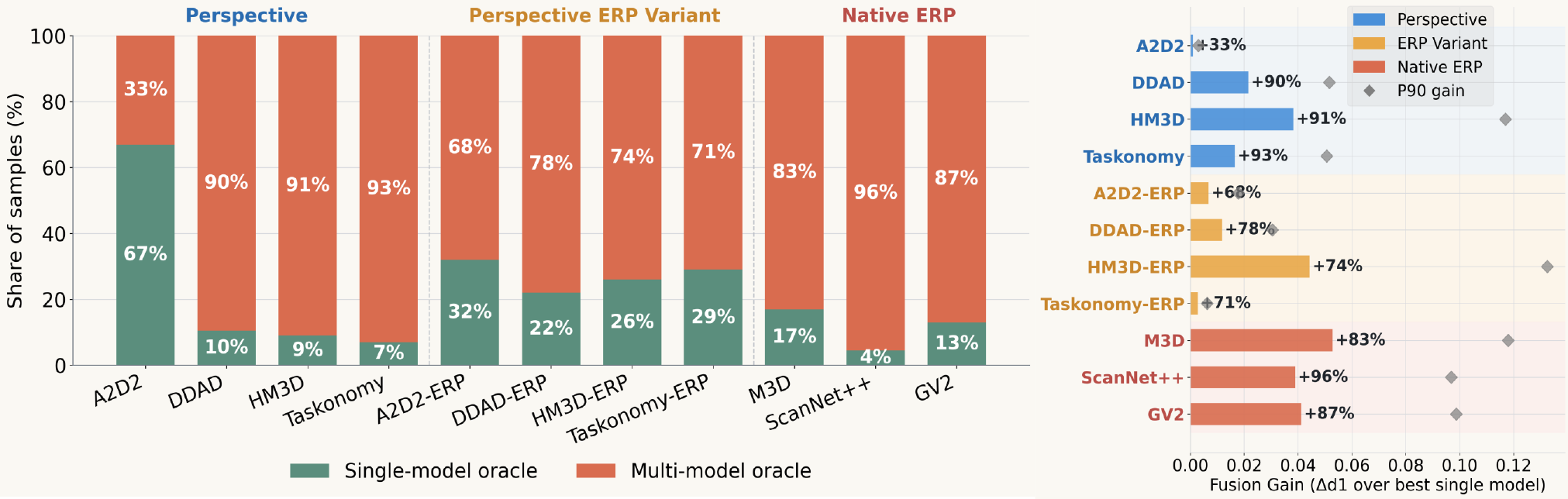}
  \caption{
    \textbf{Fusion consistently outperforms single models.}
    \textbf{Left:} Dataset-wise oracle proportions of single-model vs.\ multi-model solutions. \textbf{Right:} Fusion Gain against best single model. Bars and diamonds denote the mean and 90th percentile of per-sample $\delta_1$ fusion gain over the best single model; annotations show the fraction of samples where fusion achieves higher $\delta_1$.
  }
  \label{fig:analysis_single_fusion}
  \vspace{-1.5em}
\end{figure}

\paragraph{Fusion outperforms single models on most samples.}
In Fig.~\ref{fig:analysis_single_fusion} (left), the proportion of oracle solutions that contain multiple models could reach 96\% on \scannetpp and 93\% on \taskonomy, while the remaining samples are best handled by a single model. Across all 11 datasets, the oracle solutions are multi-model for 78.6\% of samples. Fig.~\ref{fig:analysis_single_fusion} (right) shows the average $\delta_1$ improvement of the oracle solution compared with the best single model on each dataset. Fusion improves performance to varying degrees across all datasets. In addition to the mean gain, we report the P90 gain, defined as the 90th percentile of per-sample $\Delta \delta_1$, to characterize the upper-end improvement achieved on samples where fusion is particularly beneficial. The advantage is most pronounced on native ERP datasets: fusion achieves mean gains of $\Delta \delta_1=0.053$ on \mThreeD and $0.041$ on \panoGVTwo, with 83\% and 87\%of samples benefiting from fusion, respectively.

\begin{figure}[t!]
  \centering
  \includegraphics[width=0.9\textwidth]{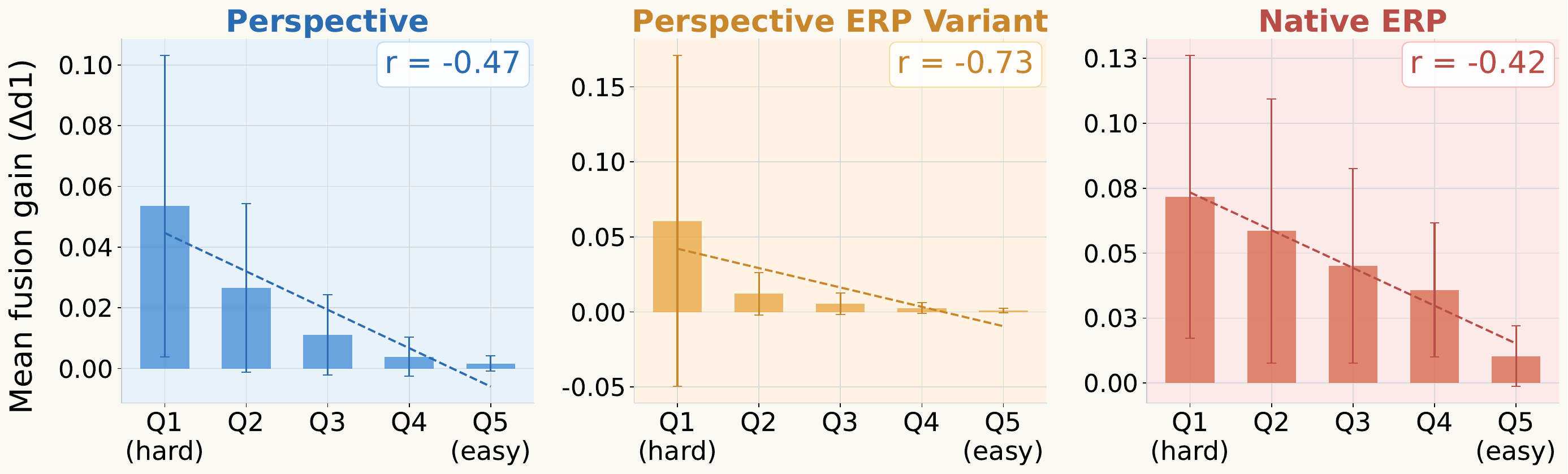}
  \caption{
    \textbf{Difficulty-dependent fusion gain.} Mean fusion gain ($\Delta \delta_1 \pm \sigma$) is shown across best-single $\delta_1$ quintiles within each dataset group, where Q1/Q5 denote the hardest/easiest samples. Fusion gains are largest on hard samples (the strongest individual model performs poorly). Dashed lines indicate linear trends over quintile means, while Pearson $r$ is computed over pooled per-sample pairs.
  }
  \label{fig:harder_more_gain}
  \vspace{-1em}
\end{figure}

\paragraph{Fusion yields larger gains on difficult samples.}
Fig.~\ref{fig:harder_more_gain} groups samples by the best single-model $\delta_1$ within each dataset family, where Q1 contains the hardest samples and Q5 the easiest. In all three settings, the largest mean fusion gains appear in the hardest quintile and steadily diminish as single-model performance improves. The negative Pearson correlations further confirm that fusion is most useful precisely when the strongest individual model is insufficient. This pattern is especially pronounced in the ERP setup, where difficult samples are more frequent, making fusion a targeted correction rather than a uniform boost. This finding suggests that \textit{fusion gains are strongly correlated with sample difficulty, rather than being uniformly distributed across samples.}

These analyses motivate \ours: the best solution is input-dependent, governed by camera geometry and sample difficulty. Thus, fusion should be a per-sample decision over when to fuse, which experts to trust, and how to combine them. We formulate this as an agent-driven process, where a VLM selects the most suitable solution for each input.

\subsection{\ours}
\begin{figure*}
  \centering
  \includegraphics[width=\textwidth]{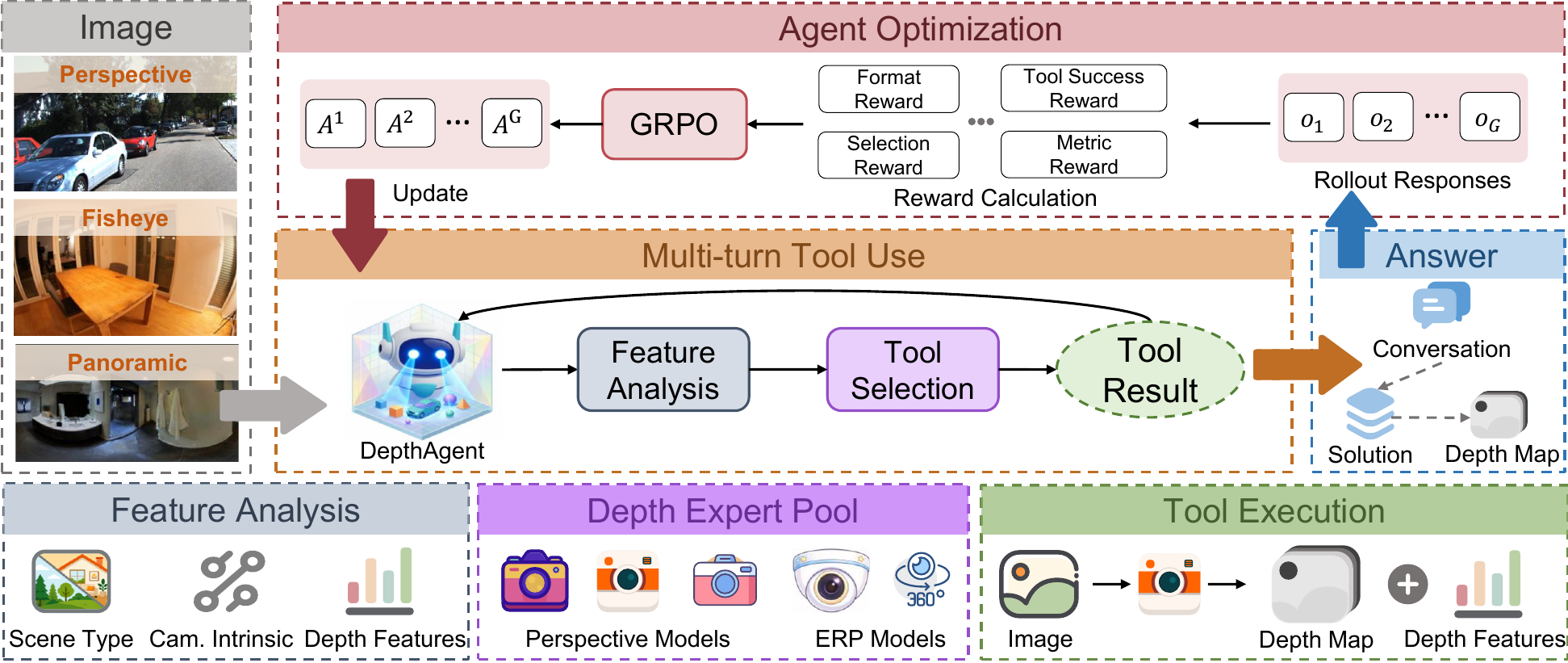}
  \caption{\textbf{Overview of \ours.} Given an input image, \ours analyzes the scene type and camera intrinsics, and interactively selects depth experts from a depth-expert tool pool. Each tool call returns a depth prediction along with auxiliary depth features. Based on the tool results, \ours determines whether to continue exploration or produce the final solution, which includes the final depth map generated by the solution. During training, rollout conversations are optimized with GRPO using the proposed multi-reward objective.}
  \label{fig:overview}
  \vspace{-1em}
\end{figure*}

The overall framework of \ours is illustrated in Fig.~\ref{fig:overview}. Given an input image $x$, \ours formulates depth estimation as an agentic process that adaptively coordinates multiple depth experts to produce the final depth solution. Since expert selection depends on high-level cues such as scene content, camera geometry, projection distortion, and intermediate prediction consistency, which are difficult to encode with a fixed hand-crafted router, \ours leverages a VLM agent to jointly reason about expert suitability and dynamically adjust tool usage for each sample.

\paragraph{Feature analysis.}
Before invoking any depth expert, the agent first considers the visual and geometric properties of the input image.
This includes scene-level cues, such as whether the image is indoor or outdoor, and camera-related cues, such as intrinsic geometry, projection characteristics, and extracted depth-related features.
These observations help the agent identify the camera type of $x$ and which depth expert is more suitable for the current sample.

\paragraph{Depth expert pool and tool selection.}
Based on the analyzed features, the agent selects depth experts from a heterogeneous depth expert pool. The pool contains both perspective models and ERP-trained models, enabling \ours to handle standard pinhole images as well as 360$^\circ$, fisheye, and other geometrically distorted inputs. At each decision step, \ours chooses an expert $a_t$ according to the current image understanding and previously observed tool outputs, instead of relying on a fixed model or a predefined fusion rule.

\paragraph{Multi-turn tool execution.}
After a successful tool call, the selected expert returns a predicted depth map $\mathbf{D}_{a_t}$ together with auxiliary depth features $f_{a_t}$, such as average depth distance and depth variation. The depth map serves as a candidate solution, while the auxiliary features provide a compact summary of the prediction quality and geometric behavior, making the tool result easier for the agent to interpret. Conditioned on the current observation and previous tool results, the agent decides whether to invoke another expert, compare complementary predictions, or terminate with a final answer. This ReAct-style~\cite{yao2023react} multi-turn process allows \ours to adaptively explore different depth experts and construct an appropriate depth solution for each sample.

\paragraph{Agent optimization.}
We use GRPO~\cite{shao2024deepseekmath} instead of SFT because fixed oracle trajectories may become invalid when the expert pool changes, and offer limited flexibility in dynamically balancing tool call efficiency and depth performance.
During training, we sample $G$ rollout responses $\{o_1, o_2, \dots, o_G\}$ for each input, where $o_i$ records the agent's reasoning, tool-use trajectory, selected solution (expert combination and fusion strategy), and corresponding final depth map. 
Each rollout is evaluated using rewards that consider both the final depth prediction and the intermediate decision process, including format correctness, tool success, selection behavior, and efficiency.
The resulting advantages $\{A^1, A^2, \dots, A^G\}$ are then used to update the agent, enabling \ours to learn how to select suitable depth tools through interaction. Furthermore, we apply multi-reward normalization~\cite{zhu2026can, liu2026gdpo} in the advantage calculation. Specifically, given $K$ reward functions ${r_{(1)}^i, r_{(2)}^i, \ldots, r_{(K)}^i}$ for the $i$-th response $o_i$, we compute the normalized advantage for each reward and aggregate as:
\begin{equation}
A_{(k)}^i = \frac{r_{(k)}^i - \mathrm{mean}({r_{(k)}^1, \ldots, r_{(k)}^G})}{\mathrm{std}({r_{(k)}^1, \ldots, r_{(k)}^G})}, A^i = \sum_{k=1}^{K} A_{(k)}^i.
\end{equation}

\subsubsection{Reward Functions} \label{subsubsec:reward_functions}

In addition to traditional rewards like format reward and tool success reward~\cite{shao2024deepseekmath, liu2025visual, zhu2026fusionagent}, we further design three reward functions: scene awareness reward $R_\text{scene}$, selection prior reward $R_\text{sel}$, and efficiency-aware metric reward $R_\text{em}$. Additional details are in the Appendix.

\paragraph{Scene awareness reward.}
Correct model selection requires the agent to understand the scene context and infer the camera geometry from the input image. We introduce a scene awareness reward to encourage the agent to first analyze whether the scene is indoor or outdoor and to estimate the camera type before selecting tools. For each sample, the Depth Agent first invokes the \texttt{estimate\_camera\_type} tool to infer whether the camera follows a perspective/pinhole or ERP/360$^\circ$ panoramic projection from camera information. This reward is implemented as a lightweight multi-label matching criterion over the first reasoning block, based on correctly identified scene category and camera type. The reward equals the average accuracy over these two labels.

\paragraph{Selection prior reward.}
The selection prior reward $R_\text{sel}$ encodes our prior knowledge about expert suitability, as observed in Sec~\ref{subsec:fusion_analysis}. It gives a mild reward when the selected experts match the input domain: perspective-oriented experts for perspective inputs, and ERP-trained experts for panoramic or large-FoV inputs. 
This provides soft guidance for faster policy adaptation in training. $R_\text{sel}$ is set to $1.0$ if the selected experts include at least one expert matching the input domain, and $0.0$ otherwise.

\paragraph{Efficiency-aware metric reward.}
Although more tool calls may improve the final prediction, the marginal gain is not always worth the additional computation. We therefore introduce an efficiency-aware metric reward $R_\text{em}$ that jointly considers the metric improvement and the number of model calls. Given a solution $y_i$ from $o_i$, let $m_i$ denote its task metric score and $n_i$ denote the number of models in the solution. $y_\text{ref}$ and $m_\text{ref}$ denote the reference solution (\eg, per-sample oracle solution) and its metric score. We define the normalized metric gap and efficiency gap:
\begin{equation}
\Delta m_{i} = \frac{m_i - m_\text{ref}}{|m_\text{ref}|+\epsilon}, \qquad
\Delta n_{i} = \frac{n_i - n_\text{ref}}{n_{\max}} .
\end{equation}
The efficiency-aware reward is then computed as
\begin{equation}
R_{\mathrm{em}}(y_i,y_\text{ref})
= \Delta m_{i}
- \lambda \Delta n_{i}\exp\left(-\frac{|\Delta m_{i}|}{\tau}\right),
\end{equation}
where $\lambda$ controls the strength of the efficiency penalty, $\tau$ controls how strongly the penalty depends on the metric gap, and $\epsilon$ is used for numerical stability. This reward discourages extra model calls when the metric improvement is small, while still allowing higher computation when it leads to a meaningful performance gain.
\section{Experiments} \label{sec:experiments}

\paragraph{Datasets.}
We train on indoor perspective and ERP variant from \hmThreeD~\cite{ramakrishnan2021habitat}, outdoor perspective data from \ddad~\cite{Guizilini2020ddad} and \aTwoDTwo~\cite{geyer2020a2d2}, and panoramic data from Matterport3D~\cite{chang2017matterport3d}. We evaluate on six diverse benchmarks across three camera domains: perspective datasets, including \kitti~\cite{Geiger2012kitti}, \nyu~\cite{silberman2012nyu}, and \ibims~\cite{koch2022ibims}; fisheye datasets \scannetpp~\cite{yeshwanth2023scannet++}; and panoramic / 360$^\circ$ datasets, including \mThreeD~\cite{chang2017matterport3d} and \panoGVTwo~\cite{albanis2021pano3d}.
We follow prior works~\cite{ganesan2026unidac, guo2025depth, piccinelli2025unik3d} to report the percentage of inliers ($\delta_i$) under thresholds of $1.25^i$, absolute relative error (\absRel), and root mean squared error (\rmse). ERP models are evaluated against ERP-transformed ground truth~\cite{ganesan2026unidac, guo2025depth}.

\paragraph{Baselines.}
We compare \ours against the \textbf{depth experts in the tool pool} and \textbf{fusion strategies} described in Sec.~\ref{subsec:fusion_analysis}. Furthermore, we compare with different selection strategies: (i) \textbf{Rand-Model:} randomly selects an expert for each sample; (ii) \textbf{Rand-Sol:} randomly selects one candidate solution. (iii) \textbf{MLP router:} an MLP as a selection router to predict the suitable solution. For Rand-Model and Rand-Sol, expert selection is restricted to perspective-specialized experts for perspective datasets, whereas selection is performed among ERP-trained candidates for other datasets. Details of the MLP router are provided in the Appendix. 

\paragraph{Implementation.}
To avoid excessive computational overhead, \ours is built on Qwen2.5-VL-3B~\cite{bai2025qwen2} and fine-tuned with GRPO~\cite{shao2024deepseekmath}. We train the agent for 300 steps with a maximum of 5 interaction turns. $\lambda$ is set to $1$ and $0.2$, and $\tau$ is set to $0.1$ and $3.4$ for perspective and ERP inputs, respectively. Training is performed on 4 H100 GPUs with an effective batch size of 4. During inference, we use greedy decoding to ensure reproducibility. Additional details are in the Appendix.

\subsection{Experimental Results}

\paragraph{Comparison with baselines.}
Tab.~\ref{tab:perspective_lfov_erp_benchmark} shows that \ours consistently outperforms individual experts and non-agentic fusion baselines across camera domains, achieving the best performance on five datasets as well as on average. Although \uniKThreeD is a strong universal-camera expert, \ours significantly improves \absRel and \rmse.  
Compared with fusion strategies, \ours shows stronger robustness, underscoring the importance of sample-wise solution selection. The comparisons with Rand-Model and Rand-Sol further indicate that agentic selection is a key contributor to the strong performance of \ours. In contrast, the MLP router shows limited routing ability and poor generalization. Notably, the improvements are most evident in \rmse, which penalizes large depth errors more heavily and is thus crucial for evaluating geometric fidelity and robustness in depth estimation. Although \ours still falls short of the upper bound, it consistently outperforms the baselines by reducing the negative impact of suboptimal fusion. Overall, these results suggest that \emph{fusion is effective only when appropriate and complementary experts are selected.}

\begin{table*}[t!]
  \centering
  \caption{
    \textbf{Quantitative comparison.}
    Perspective reports the average performance over \kitti, \nyu, and \ibims.
    Fisheye reports performance on \scannetpp, while panoramic reports the average performance over \mThreeD and \panoGVTwo.
    $^\ddagger$: best solution per sample.
  }
  \label{tab:perspective_lfov_erp_benchmark}
  \resizebox{0.92\textwidth}{!}{%
  \begin{tabular}{l|ccc|ccc|ccc}
    \toprule
    \multirow{2}{*}{\textbf{Method}}
    & \multicolumn{3}{c|}{\textbf{Perspective}}
    & \multicolumn{3}{c|}{\textbf{Fisheye}}
    & \multicolumn{3}{c}{\textbf{Panoramic}} \\
    & $\delta_1 \uparrow$ & \absRel $\downarrow$ & \rmse $\downarrow$
    & $\delta_1 \uparrow$ & \absRel $\downarrow$ & \rmse $\downarrow$
    & $\delta_1 \uparrow$ & \absRel $\downarrow$ & \rmse $\downarrow$ \\
    \midrule

    \unidepth~\cite{piccinelli2024unidepth}
    & 0.700 & 0.203 & 1.359
    & 0.135 & 0.573 & 1.029
    & 0.415 & 0.314 & 0.898 \\

    \metricThreeD~\cite{yin2023metric3d}
    & 0.850 & 0.136 & 1.279
    & 0.727 & 0.162 & 0.450
    & 0.403 & 0.290 & 0.721 \\

    \metricThreeDvTwo~\cite{hu2024metric3d}
    & 0.864 & 0.124 & 1.193
    & 0.745 & 0.170 & 0.474
    & 0.422 & 0.300 & 0.847 \\

    \unidac~\cite{ganesan2026unidac}
    & 0.845 & 0.140 & 1.876
    & 0.918 & \underline{0.097} & \underline{0.279}
    & 0.755 & 0.169 & 0.472 \\

    \uniKThreeD~\cite{piccinelli2025unik3d}
    & \underline{0.937} & \underline{0.091} & 1.261
    & \underline{0.927} & 0.103 & 0.287
    & \textbf{0.820} & \underline{0.152} & \underline{0.383} \\

    \midrule

    Mean~\cite{zhang2020ifcnn, ma2022swinfusion}
    & 0.890 & 0.110 & \underline{1.045}
    & 0.840 & 0.139 & 0.313
    & 0.620 & 0.199 & 0.529 \\

    Max~\cite{zhang2020ifcnn, ma2022swinfusion}
    & 0.590 & 0.285 & 2.276
    & 0.139 & 0.573 & 1.023
    & 0.735 & 0.235 & 0.500 \\

    Min~\cite{zhang2020ifcnn, ma2022swinfusion}
    & 0.912 & 0.095 & 1.258
    & 0.663 & 0.190 & 0.573
    & 0.266 & 0.354 & 1.091 \\

    DepthFusion~\cite{obukhov2025fourth}
    & 0.878 & 0.117 & 1.195
    & 0.621 & 0.252 & 0.623
    & 0.378 & 0.327 & 1.000 \\

    Rand-Model
    & 0.801 & 0.159 & 1.287
    & 0.921 & 0.100 & 0.284
    & 0.787 & 0.160 & 0.429 \\

    Rand-Sol
    & 0.808 & 0.150 & 1.250
    & \underline{0.927} & \underline{0.097} & \underline{0.279}
    & 0.792 & 0.158 & 0.422 \\

    MLP
    & 0.904 & 0.107 & 1.328
    & 0.187 & 1.014 & 2.547
    & 0.744 & 0.182 & 0.473 \\

    \rowcolor{green!20}
    \textbf{\ours}
    & \textbf{0.948} & \textbf{0.070} & \textbf{0.918}
    & \textbf{0.946} & \textbf{0.084} & \textbf{0.254}
    & \underline{0.819} & \textbf{0.145} & \textbf{0.365} \\

    \midrule

    Upper-bound$^\ddagger$
    & 0.985 & 0.057 & 0.844
    & 0.967 & 0.080 & 0.239
    & 0.860 & 0.132 & 0.360 \\

    \bottomrule
  \end{tabular}
  }
  \vspace{-.7em}
\end{table*}

\begin{table*}[t!]
  \centering
  \caption{
    \textbf{Comparison on hard samples.}
    We report performance on hard samples, defined as the top 10\% worst-performing samples of the best single model.
    Perspective reports the average performance over \kitti, \nyu, and \ibims.
    Fisheye reports performance on \scannetpp, while panoramic reports the average performance over \mThreeD and \panoGVTwo.
  }
  \label{tab:hard_samples_main}
  \resizebox{0.92\textwidth}{!}{%
  \begin{tabular}{l|ccc|ccc|ccc}
    \toprule
    \multirow{2}{*}{\textbf{Method}}
    & \multicolumn{3}{c|}{\textbf{Perspective}}
    & \multicolumn{3}{c|}{\textbf{Fisheye}}
    & \multicolumn{3}{c}{\textbf{Panoramic}} \\
    \cmidrule(lr){2-4}
    \cmidrule(lr){5-7}
    \cmidrule(lr){8-10}
    & $\delta_1 \uparrow$ & \absRel $\downarrow$ & \rmse $\downarrow$
    & $\delta_1 \uparrow$ & \absRel $\downarrow$ & \rmse $\downarrow$
    & $\delta_1 \uparrow$ & \absRel $\downarrow$ & \rmse $\downarrow$ \\
    \midrule

    \unidepth~\cite{piccinelli2024unidepth}
    & 0.650 & 0.271 & 2.045
    & 0.197 & 0.610 & 0.939
    & 0.337 & 0.496 & 0.853 \\

    \metricThreeD~\cite{yin2023metric3d}
    & 0.681 & 0.220 & 2.311
    & 0.635 & 0.203 & 0.629
    & 0.346 & 0.372 & 0.847 \\

    \metricThreeDvTwo~\cite{hu2024metric3d}
    & 0.737 & 0.212 & 2.072
    & 0.659 & 0.204 & 0.593
    & 0.292 & 0.427 & 0.919 \\

    \unidac~\cite{ganesan2026unidac}
    & 0.754 & 0.177 & 2.286
    & 0.789 & 0.163 & \underline{0.434}
    & 0.547 & 0.273 & 0.525 \\

    \uniKThreeD~\cite{piccinelli2025unik3d}
    & \underline{0.794} & \underline{0.155} & \underline{1.950}
    & \underline{0.799} & \underline{0.156} & \underline{0.434}
    & \underline{0.560} & \underline{0.261} & \underline{0.466} \\

    \rowcolor{green!20}
    \textbf{\ours}
    & \textbf{0.833} & \textbf{0.115} & \textbf{1.658}
    & \textbf{0.817} & \textbf{0.144} & \textbf{0.410}
    & \textbf{0.587} & \textbf{0.253} & \textbf{0.450} \\

    \bottomrule
  \end{tabular}
  }
  \vspace{-1.5em}
\end{table*}

\paragraph{Comparison on hard samples.}
Hard samples are critical because they often correspond to failure cases in real-world deployment, where improving robustness is more important than average-case performance. We further evaluate \ours on hard samples, defined as the worst 10\% samples ranked by the $\delta_1$ score of the best single model on each dataset. As shown in Tab.~\ref{tab:hard_samples_main}, \ours consistently outperforms all baseline experts on perspective, fisheye, and 360$^\circ$ data. The improvements are more pronounced than on the full test sets, especially in $\delta_1$, indicating that \ours is particularly effective at selecting reliable experts for challenging cases.

\paragraph{Efficiency of \ours.}
To improve efficiency, we introduce a Fast mode that skips CoT reasoning and directly selects tools before producing the final answer, while retaining the same selected solution as CoT mode. On an H100, \ours takes 1.1s per sample in Fast mode, close to exhaustively running all five experts at 0.76s, while CoT mode remains moderate at 3.7s. Although exhaustive fusion (\ie, Mean in Tab.~\ref{tab:perspective_lfov_erp_benchmark}) is similarly efficient, it can substantially degrade depth accuracy by indiscriminately combining unreliable predictions. 

\subsection{Ablation Studies}
\begin{table*}[t]
\centering
\small
\caption{\textbf{Ablation study of the reward design.} Combining all rewards improves overall performance, while calibrated hyperparameters drive results beyond tool usage alone. $\bar{n}$: average number of tool calls.}
\vspace{-.25em}
\label{tab:reward_ablation}
\begin{subtable}[t]{0.41\textwidth}
\centering
\caption{Leave-one-out ablation of reward components.}
\vspace{-.25em}
\label{tab:reward_component_ablation}
\resizebox{\textwidth}{!}{
\begin{tabular}{ccc|ccc}
\toprule
$R_\text{scene}$
& $R_\text{sel}$
& $R_\text{em}$
& $\delta_1 \uparrow$ 
& \absRel $\downarrow$ 
& \rmse $\downarrow$ \\
\midrule
\checkmark & \checkmark & -- & 0.832 & 0.146 & 0.827 \\
\checkmark & -- & \checkmark & 0.871 & 0.134 & 0.650  \\
-- & \checkmark & \checkmark & 0.891 & 0.102 & 0.639 \\
\checkmark & \checkmark & \checkmark & 0.905 & 0.097 & 0.623 \\
\bottomrule
\end{tabular}
}
\end{subtable}
\hfill
\begin{subtable}[t]{0.43\textwidth}
\centering
\caption{Effect of reward hyperparameters in $R_\text{em}$.}
\vspace{-.25em}
\label{tab:reward_hyperparameter_ablation}
\resizebox{\textwidth}{!}{
\begin{tabular}{ccc|ccc|c}
\toprule
$\lambda$ 
& $\tau$ 
& $n_\text{max}$
& $\delta_1 \uparrow$ 
& \absRel $\downarrow$ 
& \rmse $\downarrow$
& $\bar{n}$\\
\midrule
0.4 & 1.0 & 1.0 & 0.860 & 0.119 & 0.576 & 1.0 \\
0.1 & 2.0 & 2.0 & 0.764 & 0.166 & 0.678 & 3.2 \\
0.4 & 3.4 & 2.0 & 0.828 & 0.143 & 0.620 & 2.5 \\ 		
0.2 & 3.4 & 2.0 & 0.905 & 0.097 & 0.623 & 1.9 \\
\bottomrule
\end{tabular}
}
\end{subtable}
\vspace{-1em}
\end{table*}

\paragraph{Effects of multi-reward.}
We ablate different reward components on all datasets, as shown in Tab.~\ref{tab:reward_component_ablation}. The results suggest that relying on a partial reward is insufficient for robust solution selection. Combining $R_\text{scene}$, $R_\text{sel}$, and $R_\text{em}$ achieves the best overall performance, indicating that scene understanding, selection prior, and efficiency-aware metric optimization are complementary. 

\paragraph{Effects of $\lambda$, $\tau$, and $n_\text{max}$.}
Tab.~\ref{tab:reward_hyperparameter_ablation} studies the effects of the reward hyperparameters in $R_\text{em}$. Overall, $n_\text{max}$ mainly determines the tool-call budget and directly affects the average tool calls. 
Under a given budget, $\lambda$ modulates the penalty for tool usage, while $\tau$ controls the tolerance to metric differences among candidate solutions. This suggests that a proper $\lambda$ could balance performance and tool cost. Together, they shape the reward sensitivity and influence whether the agent favors compact or more diverse solution compositions. The results indicate that robust performance comes from a well-calibrated reward design, rather than simply increasing the number of tool calls.

\subsection{Qualitative Results}

\begin{figure*}[t!]
  \centering
  \begin{subfigure}[t]{.62\textwidth}
    \centering
    \begin{tikzpicture}
        \draw (0,0) node[inner sep=0] {\includegraphics[width=\linewidth]{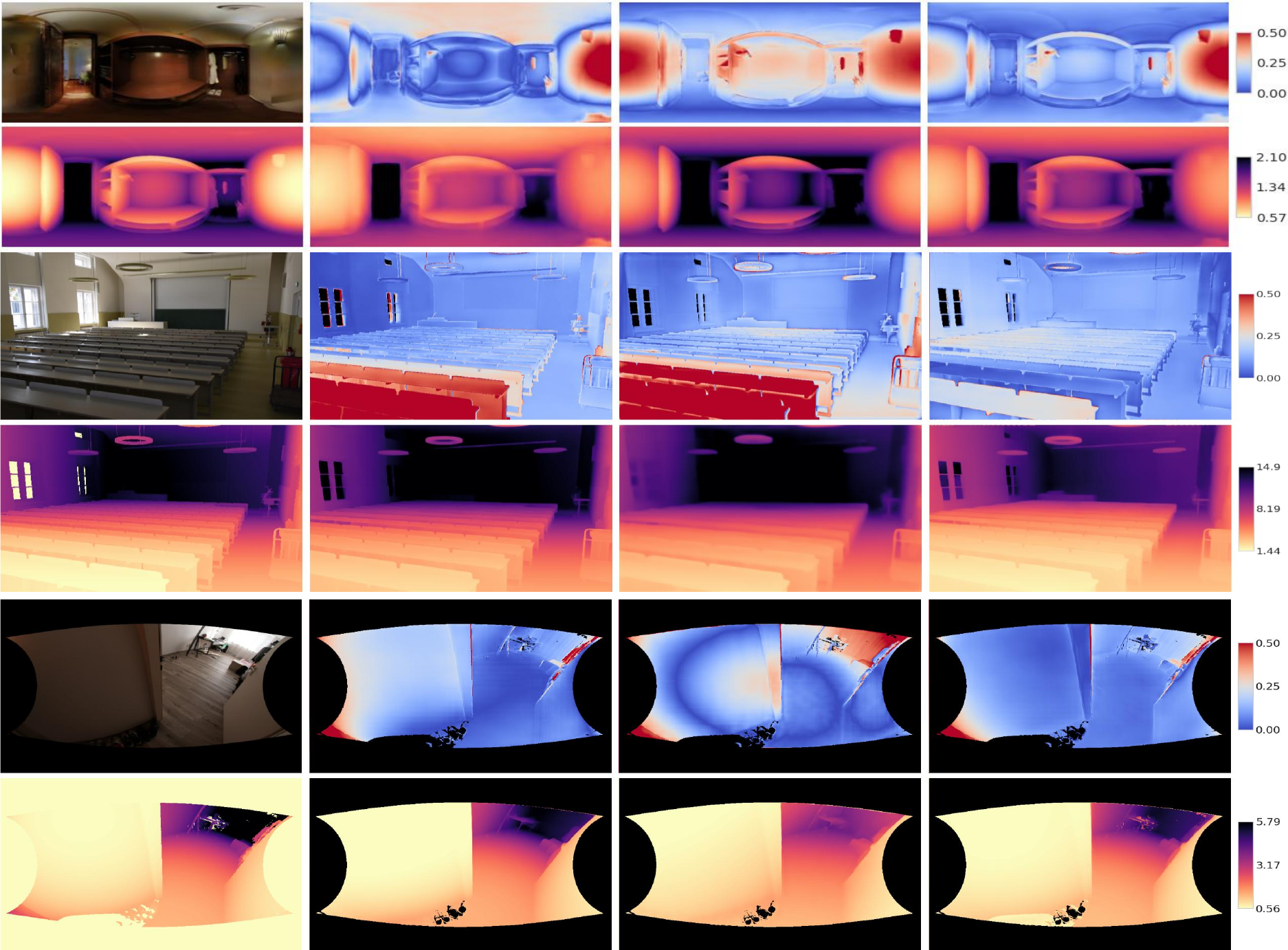}};
        \draw (-3.3,-3.4) node[inner sep=0, align=center] {\fontsize{8.0}{10}\selectfont {RGB \& GT}};
        \draw (-1.2,-3.43) node[inner sep=0, align=center] {\fontsize{8.0}{10}\selectfont {Top-1 Expert}};
        \draw (0.95,-3.43) node[inner sep=0, align=center] {\fontsize{8.0}{10}\selectfont {Top-2 Expert}};
        \draw (3.,-3.4) node[inner sep=0, align=center] {\fontsize{8.0}{10}\selectfont {Ours}};

        \draw (-4.35,3.05) node[inner sep=1, anchor=west, fill=white, fill opacity=0.5, text opacity=1] {\fontsize{4.5}{5}\selectfont {\panoGVTwo~\cite{albanis2021pano3d}}};
        \draw (-2.25,3.05) node[inner sep=1, anchor=west, fill=white, fill opacity=0.5, text opacity=1] {\fontsize{4.5}{5}\selectfont {\unidac~\cite{ganesan2026unidac}}};
        \draw (-0.22,3.05) node[inner sep=1, anchor=east, fill=white, fill opacity=0.5, text opacity=1] {\fontsize{4.5}{5}\selectfont {$\delta_1$=0.646}};
        \draw (-0.17,3.05) node[inner sep=1, anchor=west, fill=white, fill opacity=0.5, text opacity=1] {\fontsize{4.5}{5}\selectfont {\uniKThreeD~\cite{piccinelli2025unik3d}}};
        \draw (1.86,3.05) node[inner sep=1, anchor=east, fill=white, fill opacity=0.5, text opacity=1] {\fontsize{4.5}{5}\selectfont {$\delta_1$=0.632}};
        \draw (3.93,3.05) node[inner sep=1, anchor=east, fill=white, fill opacity=0.5, text opacity=1] {\fontsize{4.5}{5}\selectfont {$\delta_1$=0.856}};
        
        \draw (-4.35,1.37) node[inner sep=1, anchor=west, fill=white, fill opacity=0.5, text opacity=1] {\fontsize{4.5}{5}\selectfont {\ibims~\cite{koch2022ibims}}};
        \draw (-2.25,1.37) node[inner sep=1, anchor=west, fill=white, fill opacity=0.5, text opacity=1] {\fontsize{4.5}{5}\selectfont {\uniKThreeD~\cite{piccinelli2025unik3d}}};
        \draw (-0.22,1.37) node[inner sep=1, anchor=east, fill=white, fill opacity=0.5, text opacity=1] {\fontsize{4.5}{5}\selectfont {$\delta_1$=0.742}};
        \draw (-0.17,1.37) node[inner sep=1, anchor=west, fill=white, fill opacity=0.5, text opacity=1] {\fontsize{4.5}{5}\selectfont {\unidepth\cite{piccinelli2024unidepth}}};
        \draw (1.86,1.37) node[inner sep=1, anchor=east, fill=white, fill opacity=0.5, text opacity=1] {\fontsize{4.5}{5}\selectfont {$\delta_1$=0.697}};
        \draw (3.94,1.37) node[inner sep=1, anchor=east, fill=white, fill opacity=0.5, text opacity=1] {\fontsize{4.5}{5}\selectfont {$\delta_1$=0.884}};
        
        \draw (-4.35,-0.96) node[inner sep=1, anchor=west, fill=white, fill opacity=0.8, text opacity=1] {\fontsize{4.5}{5}\selectfont {\scannetpp~\cite{yeshwanth2023scannet++}}};
        \draw (-2.25,-0.96) node[inner sep=1, anchor=west, fill=white, fill opacity=0.8, text opacity=1] {\fontsize{4.5}{5}\selectfont {\unidac~\cite{ganesan2026unidac}}};
        \draw (-0.22,-0.96) node[inner sep=1, anchor=east, fill=white, fill opacity=0.8, text opacity=1] {\fontsize{4.5}{5}\selectfont {$\delta_1$=0.879}};
        \draw (-0.17,-0.96) node[inner sep=1, anchor=west, fill=white, fill opacity=0.8, text opacity=1] {\fontsize{4.5}{5}\selectfont {\metricThreeDvTwo~\cite{hu2024metric3d}}};
        \draw (1.86,-0.96) node[inner sep=1, anchor=east, fill=white, fill opacity=0.8, text opacity=1] {\fontsize{4.5}{5}\selectfont {$\delta_1$=0.831}};
        \draw (3.94,-0.96) node[inner sep=1, anchor=east, fill=white, fill opacity=0.8, text opacity=1] {\fontsize{4.5}{5}\selectfont {$\delta_1$=0.951}};
    \end{tikzpicture}
    
    \caption{\textbf{Depth Map Comparison.} Odd rows show the error maps, while even rows show the predicted depth maps. For each sample, we compare the top-2 expert predictions with the final solution of \ours. \ours generates more accurate depth maps with lower errors. Zoom in for better effects.}
    \label{fig:depthmap_visualization}
  \end{subfigure}
  \hfill
  \begin{subfigure}[t]{0.36\textwidth}
    \centering
    \includegraphics[width=\linewidth]{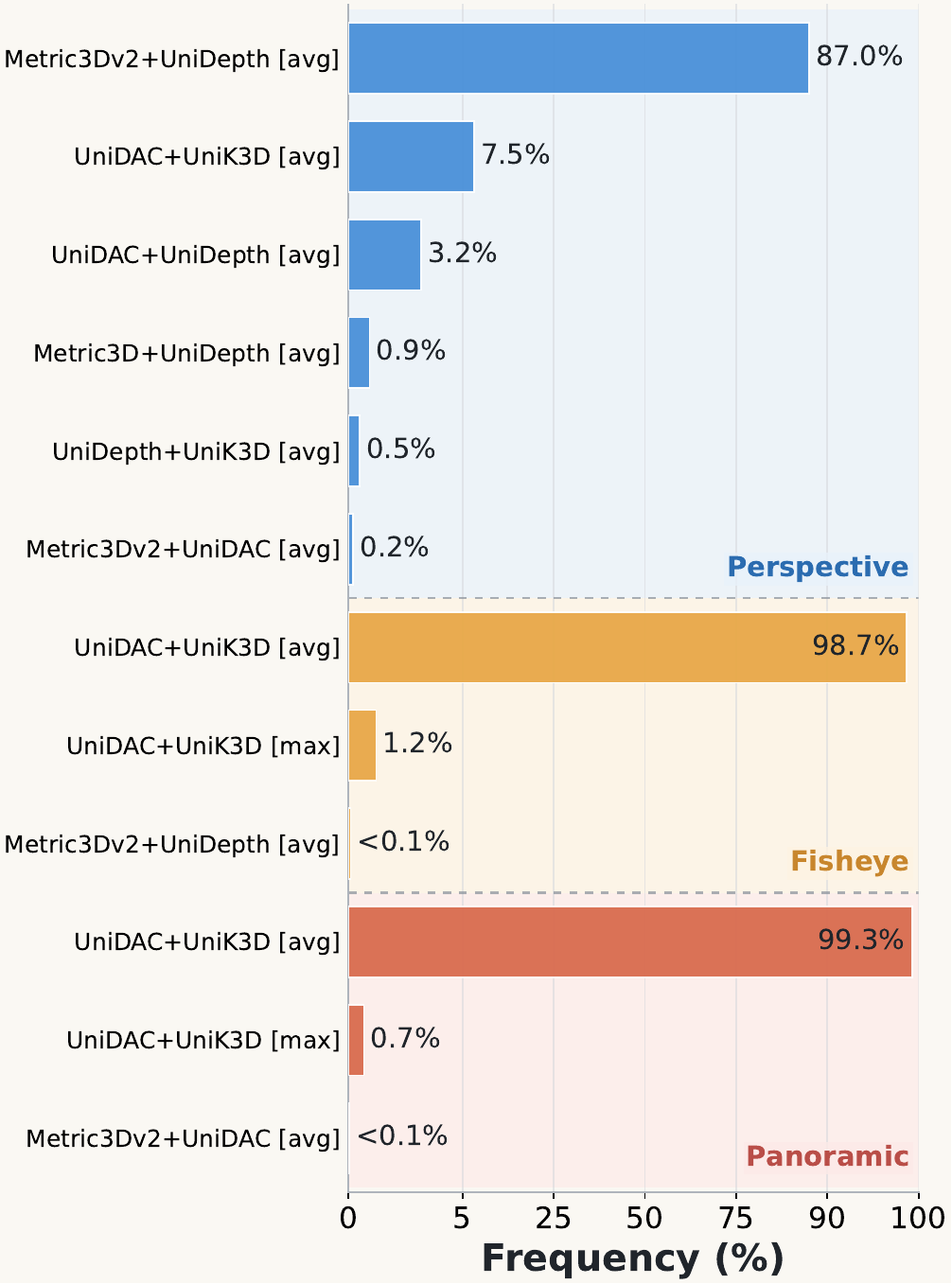}
    \caption{\textbf{Solution frequency.} Results are grouped by projection type (Perspective, Fisheye, and 360$^\circ$), and each group reports the top selected solution configurations over all test samples.}
    \label{fig:depthagent_sol_freq}
  \end{subfigure}
\caption{\textbf{Analysis of \ours behavior.}}
\vspace{-1.5em}
\end{figure*}

\paragraph{Depthmap comparison.}
Fig.~\ref{fig:depthmap_visualization} compares \ours with the top-2 individual experts on representative samples. 
\ours produces more faithful depth structures and consistently lower error maps by selecting or combining complementary expert solutions. Additional visualizations are provided in the Appendix.

\paragraph{Solution distribution on different scenarios.}
Fig.~\ref{fig:depthagent_sol_freq} shows that our agent adopts camera-dependent selection strategies under the proposed efficiency-aware metric reward. The highly concentrated distributions in fisheye and 360$^\circ$ indicate a larger metric gap between the dominant solution and other candidates, suggesting that ERP-trained experts play a leading role in these scenarios. Moreover, the policy does not simply collapse to the strongest average expert \uniKThreeD: perspective samples are often routed to \metricThreeDvTwo/\unidepth-based combinations. This also implies that appropriate expert fusion can provide stronger gains when the selected experts are well matched to the camera domain.

\paragraph{Conversation.} We provide representative visualization conversation examples in the Appendix.
\section{Conclusion} \label{sec:conclusion}

We presented \textbf{\ours}, an agentic framework for universal monocular depth estimation that performs sample-wise expert selection and fusion. Motivated by our analysis showing the benefits of fusion and camera-dependent expert preference, \ours uses a VLM agent to reason about scene and camera cues, invoke frozen depth experts, and adaptively perform the final solution. With multi-reward reinforcement fine-tuning, \ours consistently improves over baselines across perspective, fisheye, and panoramic benchmarks, especially on challenging samples. These results highlight the importance of model selection and fusion, moving beyond fixed depth estimators toward input-adaptive expert selection for systematic depth estimation.


{\small
\bibliographystyle{plainnat}
\bibliography{main}
}





\end{document}